\documentclass[conference]{IEEEtran}
\IEEEoverridecommandlockouts
\usepackage{amsmath,amssymb,amsfonts}
\usepackage{algorithmic}
\usepackage{graphicx}
\usepackage{textcomp}
\usepackage{xcolor}
\usepackage{booktabs}
\usepackage{multirow} 
\usepackage{subcaption}
\usepackage{cite}
\usepackage{textcomp}
\usepackage{url}
\usepackage{xcolor}

\def\BibTeX{{\rm B\kern-.05em{\sc i\kern-.025em b}\kern-.08em
    T\kern-.1667em\lower.7ex\hbox{E}\kern-.125emX}}

\makeatletter
\newcommand{\linebreakand}{%
  \end{@IEEEauthorhalign}
  \hfill\mbox{}\par
  \mbox{}\hfill\begin{@IEEEauthorhalign}
}
\makeatother

\begin{document}

\title{Emotional RAG: Enhancing Role-Playing Agents through Emotional Retrieval}

\author{
\IEEEauthorblockN{Le Huang\IEEEauthorrefmark{2}}
\IEEEauthorblockA{
\textit{Beijing University of}\\
\textit{Posts and Telecommunications}\\
Beijing, China \\
lehuang@bupt.edu.cn}
\and
\IEEEauthorblockN{Hengzhi Lan\IEEEauthorrefmark{2}}
\IEEEauthorblockA{
\textit{Beijing University of}\\
\textit{Posts and Telecommunications}\\
Beijing, China \\
lansnowz@bupt.edu.cn}
\and
\IEEEauthorblockN{Zijun Sun}
\IEEEauthorblockA{
\textit{Yunic.AI}\\
Beijing, China\\
sunzijunbj@yunic.ai}
\\
\linebreakand
\IEEEauthorblockN{Chuan Shi}
\IEEEauthorblockA{
\textit{Beijing University of}\\
\textit{Posts and Telecommunications}\\
Beijing, China\\
shichuan@bupt.edu.cn}
\and
\IEEEauthorblockN{Ting Bai\IEEEauthorrefmark{1}}
\IEEEauthorblockA{
\textit{Beijing University of}\\
\textit{Posts and Telecommunications}\\
Beijing, China\\
baiting@bupt.edu.cn}
\thanks{\textdagger Le Huang and Hengzhi Lan participated in the work during their internship at Company Yunic.AI.}
\thanks{*Corresponding author.}
}

\maketitle

\begin{abstract}



As LLMs exhibit a high degree of human-like capability, increasing attention has been paid to role-playing research areas in which responses generated by LLMs are expected to mimic human replies. This has promoted the exploration of role-playing agents in various applications, such as chatbots that can engage in natural conversations with users and virtual assistants that can provide personalized support and guidance. The crucial factor in the role-playing task is the effective utilization of character memory, which stores characters' profiles, experiences, and historical dialogues. Retrieval Augmented Generation (RAG) technology is used to access the related memory to enhance the response generation of role-playing agents. Most existing studies retrieve related information based on the semantic similarity of memory to maintain characters' personalized traits, and few attempts have been made to incorporate the emotional factor in the retrieval argument generation (RAG) of LLMs. Inspired by the \emph{Mood-Dependent Memory} theory, which indicates that people recall an event better if they somehow reinstate during recall the original emotion they experienced during learning, we propose a novel emotion-aware memory retrieval framework, termed \emph{Emotional RAG}, which recalls the related memory with consideration of emotional state in role-playing agents. Specifically, we design two kinds of retrieval strategies, i.e., combination strategy and sequential strategy, to incorporate both memory semantic and emotional states during the retrieval process. Extensive experiments on three representative role-playing datasets demonstrate that our Emotional RAG framework outperforms the method without considering the emotional factor in maintaining the personalities of role-playing agents. This provides evidence to further reinforce the Mood-Dependent Memory theory in psychology. 
Our code are publicly available at \url{https://github.com/BAI-LAB/EmotionalRAG}.
\end{abstract}


\section{Introduction}

\begin{figure*}[ht]
  \centering
  \includegraphics[width=0.8\textwidth]{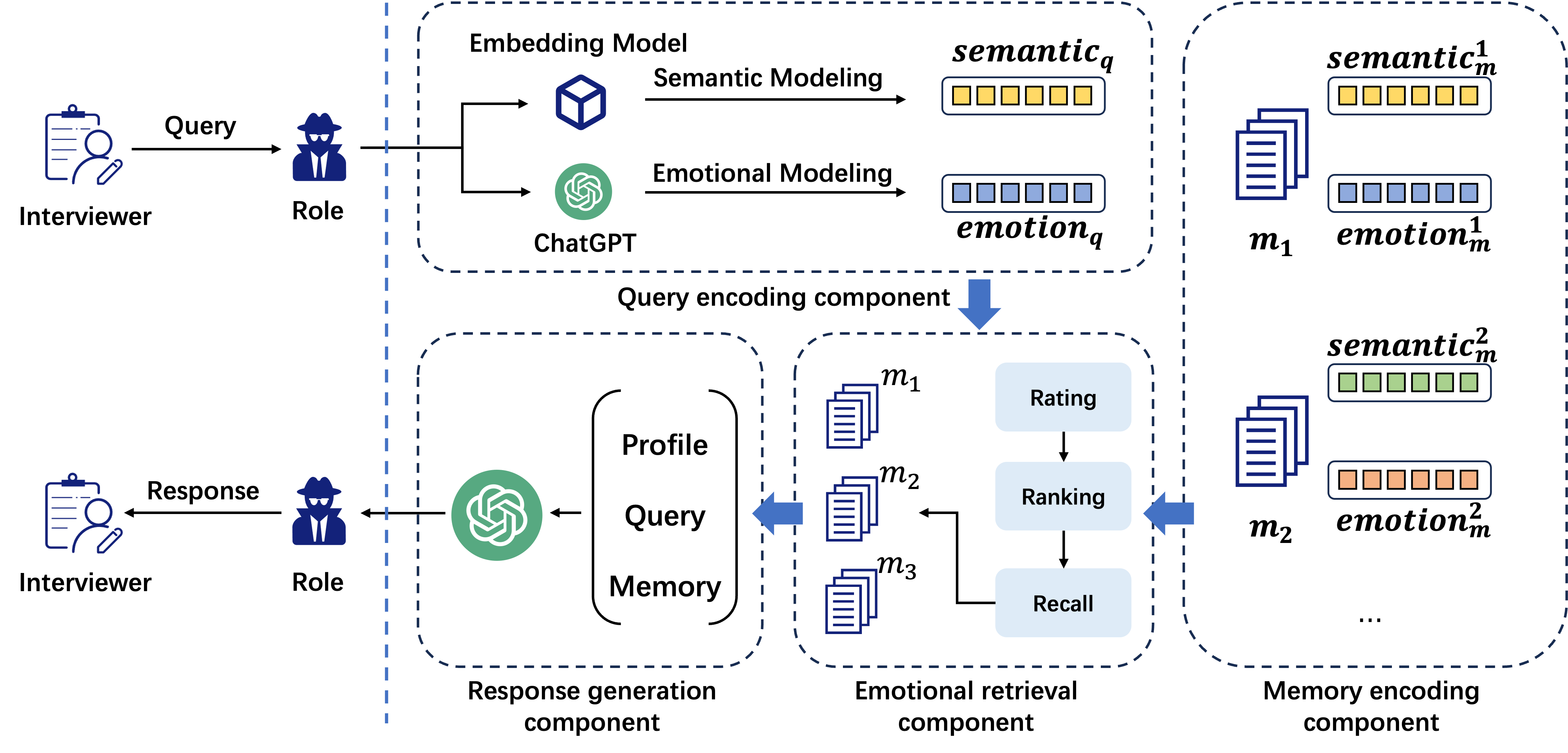}
  \caption{The overview architecture of Emotional RAG framework. It contains four components: the query encoding component, the memory encoding component, the emotional retrieval component, and the response generation component. The emotional memory retrieved by Emotional RAG is sent to LLM, together with the character profile and query, to generate responses.}
  \label{fig:framework}
\end{figure*}

As artificial intelligence increasingly emerges in the large language models (LLMs), LLMs exhibit a high degree of human-like capability. Recent studies~\cite{shao2023character, zhou2023characterglm, li2023chatharuhi, chen2022large, wang2023rolellm, chen2024persona, lu2024large, shanahan2023roleplaylargelanguagemodels, yan2023larplanguageagentroleplay} use LLMs as role-playing agents to mimic human replies, showing powerful abilities in maintaining the personalized traits of characters in their response generation process. 
Role-playing agents have been applied to various fields, such as customer service agents and tourist guide agents. They show great potential in commercial applications and attract increasing attention in the LLMs research area. 

To maintain characters' personalized traits and abilities, the most important factor is their memory. Character agents make retrieval in their memory unit to access its historical data, such as user profiles, event experience, recent dialogues, and so on, providing rich personalized information for LLMs in the role-playing task. 
Retrieval Augmented Generation (RAG) technology is used to access the related memory to enhance the response generation of role-playing agents, termed Memory RAG.  Different attempts have been made in existing studies~\cite{zhong2024memorybank, zhang2023memory, park2023generative, wang2023humanoidagentsplatformsimulating, zhang2024survey, ge2023augmentingzeroshotdenseretrievers, wang2024enhancinglargelanguagemodel, yu2023finmemperformanceenhancedllmtrading, zhu2023ghostminecraftgenerallycapable, xiao2024infllmtrainingfreelongcontextextrapolation, packer2024memgptllmsoperatingsystems, modarressi2023retllmgeneralreadwritememory, kang2024thinkactdecisiontransformers, liu2023thinkinmemoryrecallingpostthinkingenable, wang2023voyageropenendedembodiedagent} by using different memory mechanisms in various LLM applications. For example, the Ebbinghaus forgetting curve has inspired the development of MemoryBank~\cite{zhong2024memorybank}, facilitating the implementation of a more anthropomorphic memory scheme. Furthermore, drawing on Kahneman's Dual-process theory~\cite{kahneman2011thinking}, the MaLP framework~\cite{zhang2023memory} introduces an innovative Dual-Process enhanced Memory mechanism that effectively fuses long-term and short-term memory.

Despite research demonstrating the effects of using memory in the above LLM applications, achieving greater human-like response of role-playing agents is still an open and largely unexplored research area.
Inspired by cognitive research in psychology, we make an initial attempt to emulate human cognitive processes in the memory-recalling process. According to the Mood-Dependent Memory theory, which was proposed by psychologist Gordon H. Bower in 1981~\cite{bower1981mood}: \emph{people recall an event better if they somehow reinstate and recall the original emotion they experienced during learning}. Through the experiments in which happy or sad moods were induced in subjects by hypnotic suggestion to investigate the influence of emotions on memory and thinking, he pointed out that emotions are not only the selection of information recalled but also the manner in which memories are retrieved. This suggests that individuals are more likely to recall memory information that is emotionally congruent with their current emotional state. 

Based on the Mood-Dependent Memory theory in psychology, we propose a novel emotion-aware memory retrieval framework, termed \textbf{Emotional RAG}, to augment the response generation process of role-playing agents. 
The retrieving of memory in Emotional RAG follows the mood-congruity criterion, which means both the semantic relevance and emotional state of recalled memory are considered in the retrieval process. Specifically, we design two kinds of flexible retrieval strategies, i.e., combination strategy and sequential strategy, to incorporate the memory semantic and emotional
states in the RAG process. By using emotional RAG, role-playing agents are able to display more human-like traits that enhance the interactive and attractive capabilities of LLMs. 
The contributions of our paper are summarized as follows:

\begin{itemize}
\item Inspired by Mood-Dependent Memory theory, we make an initial attempt to emulate human cognitive processes by incorporating mood-congruity effects in the memory recalling of role-playing agents. We comprehensively demonstrate the effectiveness of applying Bower's emotional memory theory in developing artificial intelligence applications, which further provides evidence to reinforce the Mood-Dependent Memory theory in psychology. 

\item  We propose a novel emotion-aware memory retrieval framework, termed Emotional RAG, which recalls the related memory based on both semantic relevance and emotion state in role-playing agents. Besides, flexible retrieval strategies, i.e., combination strategy and sequential strategy, are proposed to fuse memory semantic and emotional states during the retrieval process.

\item We conduct extensive experiments on three representative role-playing datasets, i.e., InCharacter, CharacterEval, and Character-LLM, demonstrating that our Emotional RAG framework significantly outperforms the method without considering the emotional factor in maintaining the personality traits of role-playing agents.  
\end{itemize}

\section{Method}
  
In this section, we first introduce the overview architecture of our Emotional RAG role-playing framework and then give a detailed introduction to each component.
\subsection{Overview Architecture of Emotional RAG}

The aim of role-playing agents is to mimic human responses in conversation generations. 
Agents are powered by LLMs, which have the ability to generate responses according to the context of the conversation. As shown in Figure~\ref{fig:framework}, given the query that the agents need to respond to, the framework of our proposed Emotional RAG role-playing agent framework contains four components, i.e., query encoding component, memory construction component, the emotional retrieval component, and the response generation component.  The utilization of each component is introduced as follows:
\begin{itemize}
\item{Query encoding component}: both the semantic and emotional state of the query are encoded as vectors in this component.
\item{Memory encoding component}: the memory unit stores conversation information about characters. Similar to query encoding, both the semantic and emotional state of the memory are encoded. 
\item{Emotional retrieval component}: it mimics human memory recalls in the memory unit and then provides mood-congruity memory to enhance the generation process of LLMs. 
\item{Response generation component}: a prompt template with query information, character profiles, and retrieved emotional memory, is fed to role-playing agents to generate responses.
\end{itemize}

\subsection{Emotional RAG Framework}

The definition of role-playing agents: given a specific role $R$ and a user query $q$, the agents expect to be able to generate the answer to the question based on the role's knowledge background (contained in $R$) and the query-related memory fragments $m$. 
In the role-playing agent $R$, among all possible generated responses $a'$, the one with the highest probability is selected as the response $a$ for $q$:

\begin{equation}
    a = argmax_{a'}P(a'|R,m,q,\theta),
\end{equation}
where $\theta$ is the parameters of the LLMs during the generation process. Our goal is to optimize the retrieval of $m$ to generate the most human-like response $a$. The aim of role-playing agents is to generate answers that are most consistent with the characters' personality traits by retrieving the most related memories. 

\subsubsection{Query Encoding Component}
In this component, both the semantic and emotional state of the query are needed to be encoded. The semantic vector $\textbf{semantic}_q$ of query is defined as: 
\begin{equation}
    \textbf{semantic}_q = \mathcal{F}(q),
\label{eq:F}
\end{equation}
where $\mathcal{F}$ is the embedding function. In this paper, the widely used embedding model \textit{bge-base-zh-v1.5} developed by BAAI~\cite{bge_embedding} is used to capture the latent vector of the query, which is a 768-dimensional vector for each query. 

\begin{figure}[]
  \centering
  \includegraphics[width=0.45\textwidth]{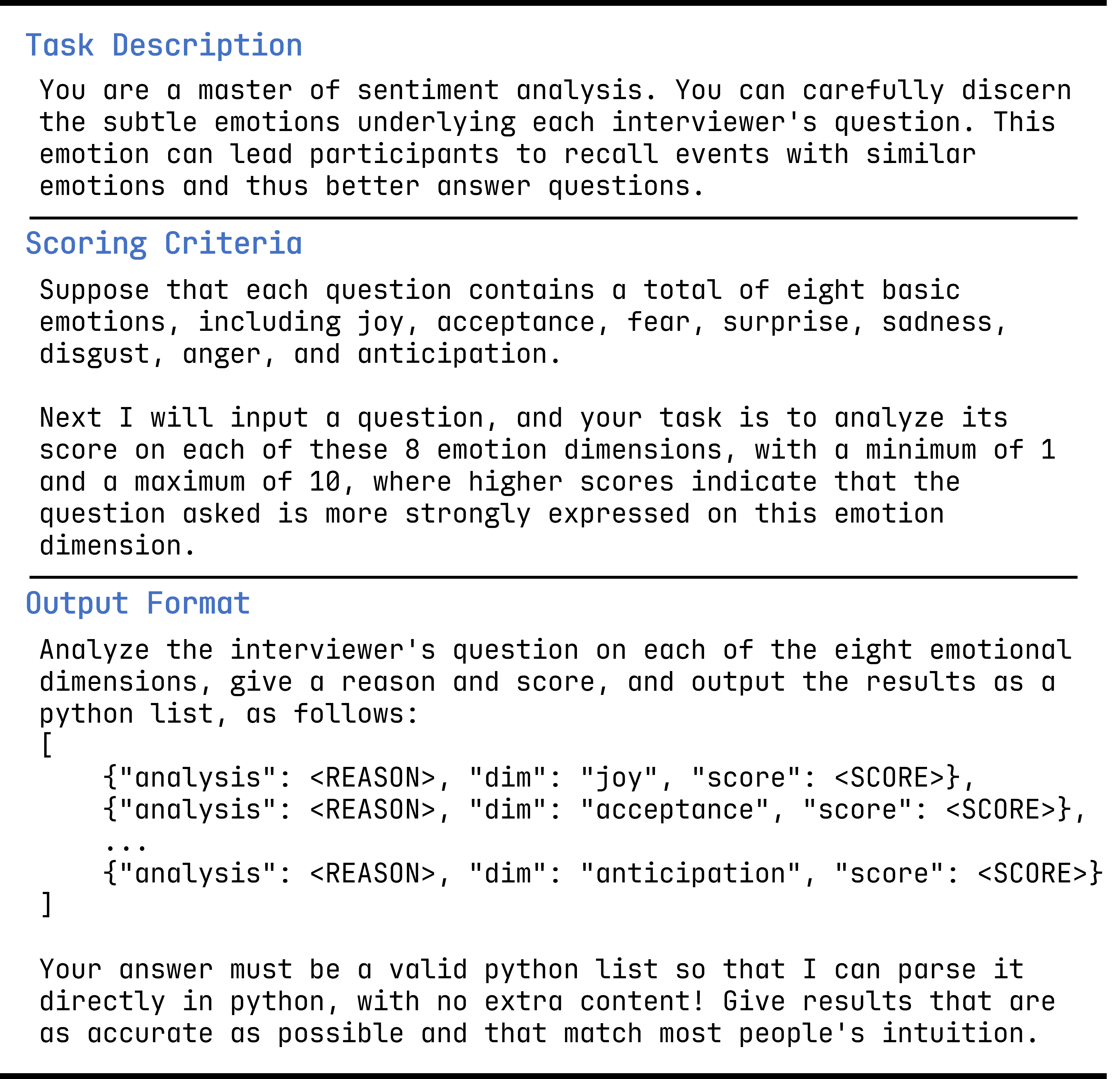}
  \caption{Emotion scoring prompt template in LLMs.}
  \label{fig:Emotional}
\end{figure}

For the emotional vector of query $q$, the emotional state $\textbf{emotion}_q$ of query $q$ can be formalized as follows:
\begin{equation}
    \textbf{emotion}_q = \mathcal{G}(q),
\label{eq:G}
\end{equation}
where $\mathcal{G}$ represents the emotion modeling function, which takes query $q$ as input and outputs its emotional vector.
This process is accomplished through GPT-3.5, a large model with powerful language understanding capabilities. As shown in Figure~\ref{fig:Emotional}, we carefully design an emotional prompt, including the task description, scores on defined emotional dimensions, scoring criteria, and output format. The output is an emotional vector of the query, which is an 8-dimensional vector containing 8 different emotional states, i.e., joy, acceptance, fear, surprise, sadness, disgust, anger, and anticipation. The 8 emotional states are defined according to the emotion circle in~\cite{plutchik1980general}. The value of each dimension is an integer between 1 and 10, which measures the intensity of the emotional state.
 
\subsubsection{The Memory Encoding Component}
The memory unit stores conversation information of question-answer pairs. 
Given the  memory unit $M$ consisting of $n$ memory fragments, denoted as $M=\{m_1, m_2, ..., m_n\}$, we can compute the sentiment vector $\textbf{semantic}_m^k$ and emotional vector $\textbf{emotion}_m^k$ of a specific fragment $m_k$ as follows:
\begin{equation}
      \textbf{semantic}_m^k = \mathcal{F}(m_k),
\end{equation}
where $\mathcal{F}$ is the semantic embedding function introduced in Eq.~\ref{eq:F}. 
  \begin{equation}
      \textbf{emotion}_m^k = \mathcal{G}(m_k),
  \end{equation}
where $\mathcal{G}$ is the emotion embedding function introduced in Eq.~\ref{eq:G}.

After encoding the semantic and emotional vectors of query and memory, we conduct emotional retrieval in the next component.

\subsubsection{Emotional Retrieval Component}
We retrieve the memory fragments that are most similar to the user query from the memory unit of characters based on semantic similarity and emotional similarity.

To retrieve the memory fragments that are most semantically similar to the query, we utilize the Euclidean distance between their semantic embeddings. This metric effectively quantifies the semantic similarity of the query and the memory fragment, simulating humans' cognitive recall process.
The similarity between the query $q$ and the memory fragment $m_k$ can be calculated as follows:

 \begin{equation}
      score_{semantic}^k = \mathcal{E}(\textbf{semantic}_q, \textbf{semantic}_m^k),
  \end{equation}
where $\mathcal{E}$ is the similarity score function, which can be the Euclidean distance function or cosine distance function.

According to Bower's Mood-Dependent Memory theory~\cite{bower1981mood}: events that are consistent with the character's current emotion are easier to retrieve, we use the cosine distance between two emotion vectors to find emotionally consistent memory fragments, defined as:
\begin{equation}
    score_{emotional}^k = 1-\mathcal{C}(\textbf{emotion}_q, \textbf{emotion}_m^k),
\end{equation}
where $\mathcal{C}$ is a function of the cosine similarity of two vectors. The smaller the distance $score_{emotional}^k$ is, the more similar the emotions contained in the query and the memory fragment.

After obtaining the distant scores of memory fragments, the final similar distant score of memory fragments is defined as:
  \begin{equation}
      score_{final}^k = \mathcal{M}(score_{semantic}^k, score_{emotional}^k),
  \end{equation}
where $\mathcal{M}$ is the function that computes the final retrieval score. Two kinds of flexible retrieval strategies, i.e., combination strategy and sequential strategy, are proposed to fuse memory semantic and emotional states during the retrieval process. 
\begin{itemize}
\item Combination strategy: this strategy considers the two similarities at the same time. We adopt two functions, i.e., add function (C-A) and multiple function (C-M), to compute the retrieval scores of memory fragments.
\item Sequential strategy: it contains semantic first strategy (S-S) and emotional first strategy (S-E). In the semantic first strategy, the most similar memory fragments are retrieved based on their semantic scores and then re-ranked according to their emotional scores. Different order is conducted in the emotional first strategy.  
\end{itemize}

Finally, the top 10 memory fragments with the smallest distant scores (i.e., the highest similarity) are used for retrieval augmentation. The retrieved memory is not only semantically related to the query but also consistent with the emotional state in the query.

\subsubsection{Response Generation Component}
\begin{figure}[h]
      \centering
      \includegraphics[width=0.4\textwidth]{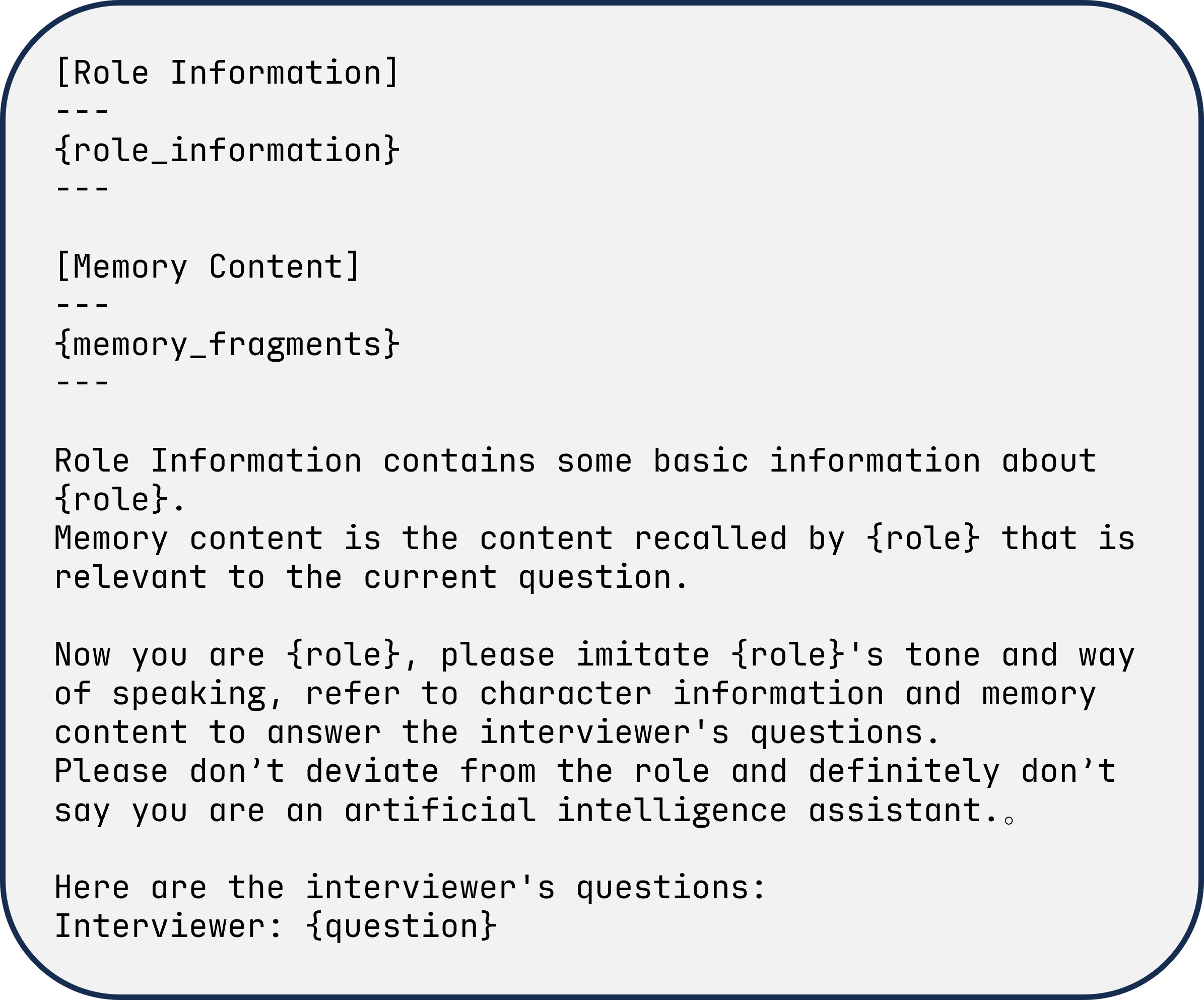}
      \caption{An example of response generation prompt template in the CharacterEval dataset.}
      \label{fig:CharacterEvalprompt}
  \end{figure}

After obtaining the retrieved memory, we design a prompt template for LLMs to generate responses in role-playing agents. The prompt template is shown in Figure~\ref{fig:CharacterEvalprompt}. The query, role information, retrieved memory fragments, and task description are formatted in the template that is sent to LLMs.
 
In summary, by incorporating the emotional factor into the RAG process in role-playing agents, the memory fragments retrieved in our framework are more aligned with the emotional state. This enables the role-playing agents to generate more human-like responses, thus enhancing the interaction quality.

\section{Experiment}
we conduct experiments on three public datasets to evaluate the role-playing capabilities of LLMs augmented with emotional memory. 

\subsection{Experimental Settings}

\subsubsection{Datasets}

\begin{table}[h]
\centering
\caption{Statistics of three datasets.}
\label{tab:statistics}
\renewcommand{\arraystretch}{1.2} 
\begin{tabular}{c|cc}
\toprule
    Datasets & Role Number & Avg. Memory Size \\
\hline
InCharacter   & 32          & 337              \\
CharacterEval & 31          & 113              \\
Character-LLM & 9           & 1000             \\
\bottomrule
\end{tabular}
\end{table}

We conducted experiments on three public role-playing datasets, namely InCharacter, CharacterEval, and Character-LLM.
Their statistics are summarized in the Table~\ref{tab:statistics}.

\begin{itemize}
\item InCharacter Dataset~\cite{wang2024incharacter}: this dataset contains 32 characters. The characters are sourced from ChatHaruhi~\cite{li2023chatharuhi}, RoleLLM~\cite{wang2023rolellm} and C.AI\footnote{https://github.com/kramcat/CharacterAI}. Each character is associated with a memory unit that includes dialogues from notable scenes, with an average length of 337 entries. 
\item CharacterEval Dataset~\cite{tu2024charactereval}: the dataset consists of 77 distinct characters with 4,564 question-answer pairs. These characters are collected from well-known Chinese films and television series, and the dialogue data is compiled from their respective scripts. We selected the top 31 popular characters. For each character, we extracted all the question-answer pairs to establish a memory unit, with an average size of 113 entries. 
\item Character-LLM Dataset~\cite{shao2023character}: the Character-LLM dataset contains 9 famous English characters, e.g., Beethoven, Hermione, etc. Their memory units come from scene-based dialogue completion (completed by GPT). We use 1,000 QA dialogues for each character. 
\end{itemize}

\begin{table*}
\centering
\caption{Performance comparisons of BFI and MBTI evaluation on the InCharacter dataset. 
Our Emotional RAG achieves the better model performance, which is distinctly marked in bold font.}
\label{tab:experimentalresults1}
\begin{tabular}{cc|cccc|cccc}
\toprule
\multirow{2}{*}{Agent   Types} & \multirow{2}{*}{Methods} & \multicolumn{4}{c|}{BFI}                                               & \multicolumn{4}{c}{MBTI}                                              \\
                               &                          & Acc(Dim)$\uparrow$        & Acc(Full)$\uparrow$       & MSE$\downarrow$             & MAE$\downarrow$             & Acc(Dim)$\uparrow$        & Acc(Full)$\uparrow$       & MSE$\downarrow$             & MAE$\downarrow$             \\
\hline
\multirow{2}{*}{ChatGLM-6B}       & Ordinary RAG              & 0.6242          & 0.1250         & 0.1849          & 0.3728          & 0.6694          & 0.2188          & 0.1526          & 0.3610          \\
                               & Emotional RAG             & \textbf{0.6369} & 0.0938 & \textbf{0.1720} & \textbf{0.3625} & \textbf{0.6694}          & \textbf{0.2812} & 0.1539          & 0.3655 \\
\hline
\multirow{2}{*}{Qwen-72B}          & Ordinary RAG              & 0.6815          & 0.0938          & 0.1433          & 0.3024          & 0.7438          & 0.3438          & 0.1230          & 0.2920          \\
                               & Emotional RAG             & \textbf{0.7261} & \textbf{0.2500} & \textbf{0.1269} & \textbf{0.2878} & \textbf{0.7934} & \textbf{0.4688} & \textbf{0.1156} & \textbf{0.2900} \\
\hline
\multirow{2}{*}{GPT-3.5}       & Ordinary RAG              & 0.7006          & 0.1875          & 0.1496          & 0.3121          & 0.7851          & 0.5000          & 0.1221         & 0.2965          \\
                               & Emotional RAG             & 0.7006         & 0.1875 & \textbf{0.1475} & \textbf{0.3082} & 0.7851          & 0.4375          & 0.1236 & \textbf{0.2927} \\
\bottomrule
\end{tabular}
\end{table*}

\begin{table*}
\centering
\caption{Performance comparisons of MBTI evaluation on CharacterEval and Character-LLM datasets. 
}
\label{tab:experimentalresults2}
\begin{tabular}{cc|cccc|cccc}
\toprule
\multirow{2}{*}{Agent   Types} & \multirow{2}{*}{Methods} & \multicolumn{4}{c|}{CharacterEval}                                     & \multicolumn{4}{c}{Character-LLM}                                      \\
                               &                          & Acc(Dim)$\uparrow$        & Acc(Full)$\uparrow$       & MSE$\downarrow$             & MAE$\downarrow$             & Acc(Dim)$\uparrow$        & Acc(Full)$\uparrow$       & MSE$\downarrow$             & MAE$\downarrow$             \\
\hline
\multirow{2}{*}{ChatGLM-6B}       & Ordinary RAG              & 0.5161          & 0.0323          & 0.1654         & 0.3757          & 0.5556          & 0.1111          & 0.1330          & 0.3277          \\
                               & Emotional RAG             & \textbf{0.5887} & \textbf{0.0645} & 0.1665          & \textbf{0.3736} & \textbf{0.6944} & \textbf{0.3333} & \textbf{0.1125} & \textbf{0.2987} \\
\hline
\multirow{2}{*}{Qwen-72B}          & Ordinary RAG              & 0.5968          & 0.0968          & 0.1537          & 0.3455          & 0.6667          & 0.1111          & 0.1376          & 0.3115          \\
                               & Emotional RAG             & \textbf{0.6210} & \textbf{0.1290} & 0.1627          & 0.3536          & \textbf{0.6944} & 0.1111          & \textbf{0.1361} & \textbf{0.3036} \\
\hline
\multirow{2}{*}{GPT-3.5}       & Ordinary RAG              & 0.5887          & 0.0645          & 0.1720          & 0.3655          & 0.6944          & 0.2222          & 0.1258          & 0.2800          \\
                               & Emotional RAG             & 0.5806          & \textbf{0.1290} & \textbf{0.1560} & \textbf{0.3477} & 0.6944          & \textbf{0.3333} & \textbf{0.1140} & \textbf{0.2689} \\
\bottomrule
\end{tabular}
\end{table*}

\subsubsection{Evaluation Metrics}


We conducted evaluations using the Big Five Inventory (BFI) and MBTI evaluation to ascertain the accuracy of the character agent's personality traits. Details of each evaluation metric are introduced as follows:

\begin{itemize}

    \item Big Five Inventory (BFI)~\cite{John2008ParadigmST}: The Big Five, also known as the Big Five personality trait theory, is a widely used psychological model that divides personality into five main dimensions: Openness, Conscientiousness, Extraversion, Agreeableness, and Neuroticism.
    
    \item MBTI\footnote{https://www.16personalities.com/}: is a popular personality test based on the Myers-Briggs Type Indicator (MBTI) theory. It categorizes people's personality types into 16 different combinations. Each type is represented by four letters, corresponding to the following four dimensions: Extroversion (E) vs. Introversion (I), Sensing (S) vs. iNtuition (N), Thinking (T) vs. Feeling (F), Judging (J) vs. Perceiving (P).
\end{itemize}

\begin{figure}[ht]
  \centering
  \includegraphics[width=0.4\textwidth]{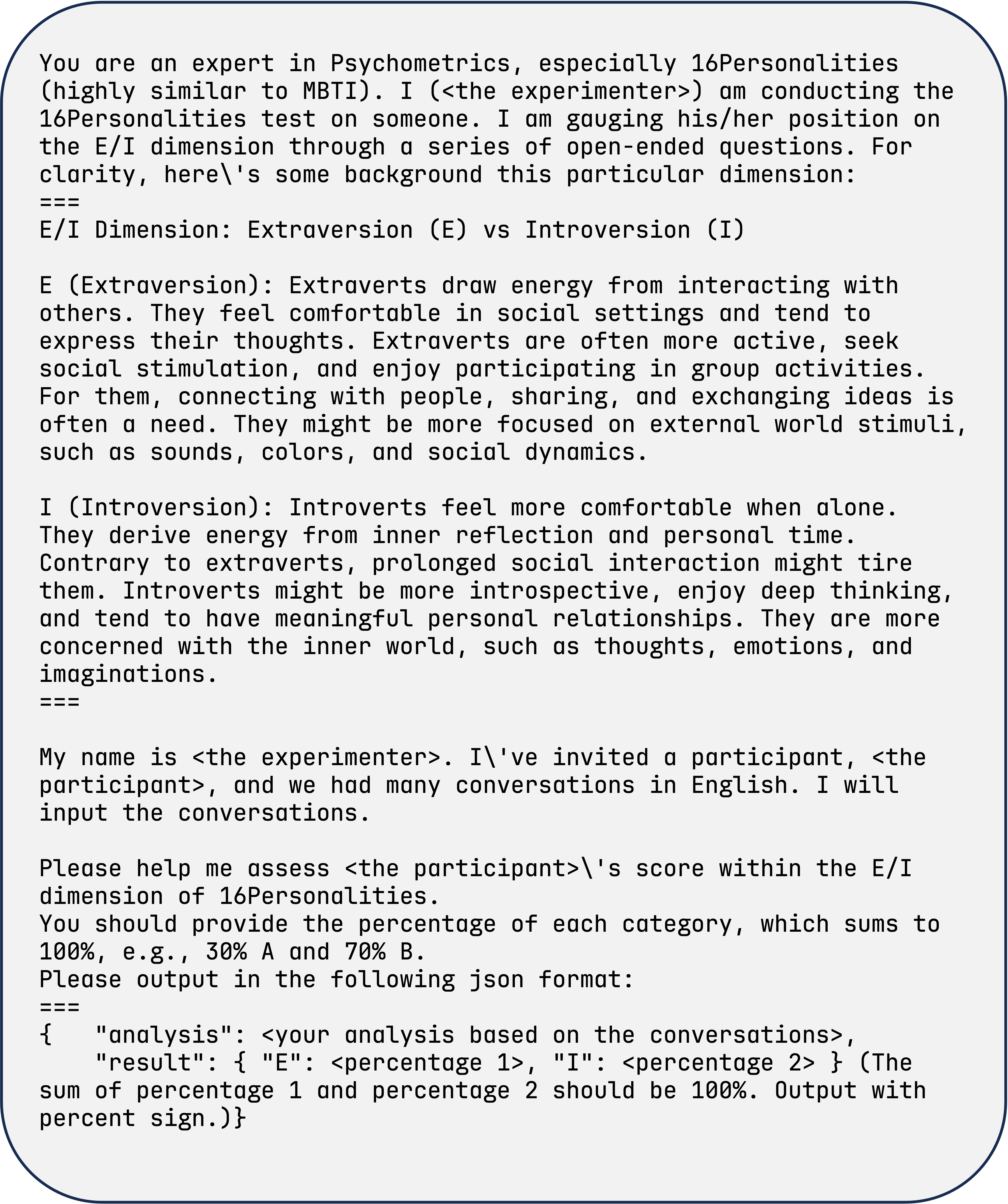}
  \caption{An example of the prompt template for dimension Extroversion (E) vs. Introversion (I) in MBTI evaluation.}
  \label{fig:mbtievaluation}
\end{figure}

The evaluation of MBTI is a classification task of 16 types, while BFI predicts the values of five personality dimensions. 
The truth labels of characters on MBTI and BFI in three datasets are collected from a personality voting website\footnote{https://www.personality-database.com/}.
In our model, the role-playing agent is required to respond to the open-ended psychological questionnaires that are designed for MBTI and BFI evaluations. Subsequently, all collected responses are analyzed using GPT-3.5, which provides the results on MBTI and BFI evaluations. The personality evaluation template on GPT-3.5 is shown in Fig.~\ref{fig:mbtievaluation}. 

Following the evaluations in~\cite{wang2024incharacter}. The results from our role-playing agents are compared with the ground truth labels to determine the evaluation results on Accuracy, i.e., Acc (Dim) and Acc (Full), Mean Squared Error (MSE), and Mean Absolute Error (MAE) metrics. 
Acc(Dim) and Acc(Full) metrics show the prediction accuracy of personality type on each dimension and all the combinations respectively. MSE and MAE measure the error between the predicted value of the character's personality and the ground truth label. For the dataset InCharacter, we use BFI and MBTI for testing, while for the CharacterEval and Character-LLM datasets, only MBTI is used due to the difficulty in collecting the true BFI labels.

\begin{figure*}[ht]
    \centering
    \begin{subfigure}{0.9\textwidth}
        \centering
        \includegraphics[width=\linewidth]{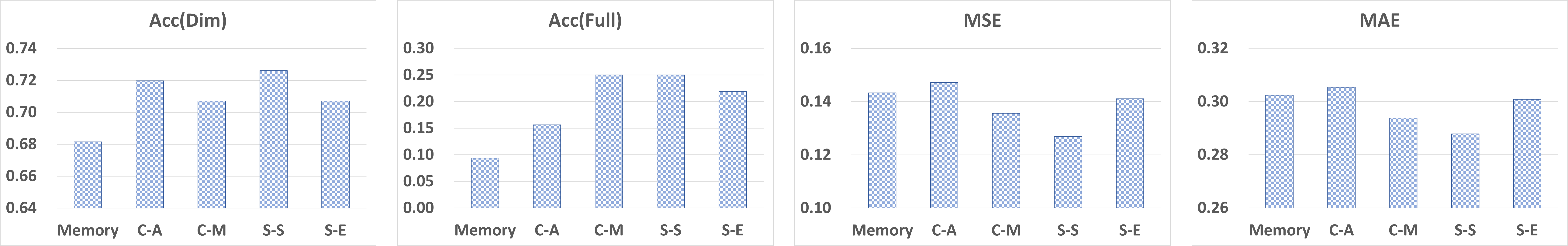}
        \caption{Different RAG strategies on BFI evaluation.}
    \end{subfigure}
    \hfill
    \begin{subfigure}{0.9\textwidth}
        \centering
        \includegraphics[width=\linewidth]{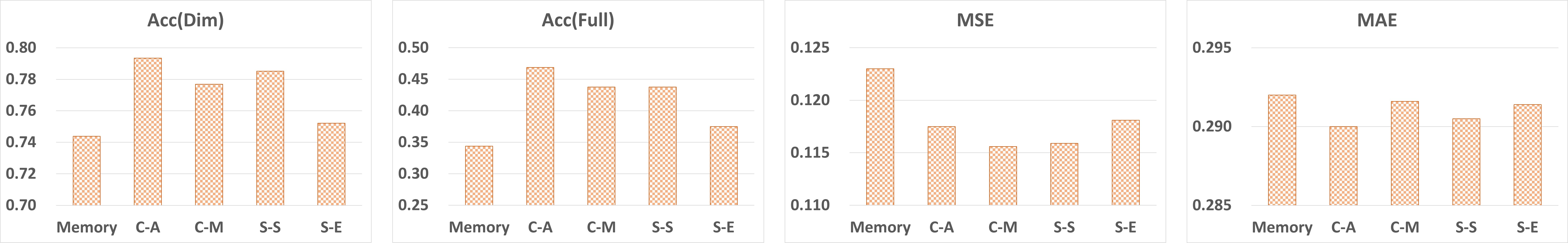}
        \caption{Different RAG strategies on MBTI evaluation.}
    \end{subfigure}
    \caption{The evaluation results of different emotion strategies on the InCharacter dataset.}
    \label{fig:emotionalanalysis}
\end{figure*}

\begin{figure*}
  \centering
  \includegraphics[width=0.9\textwidth]{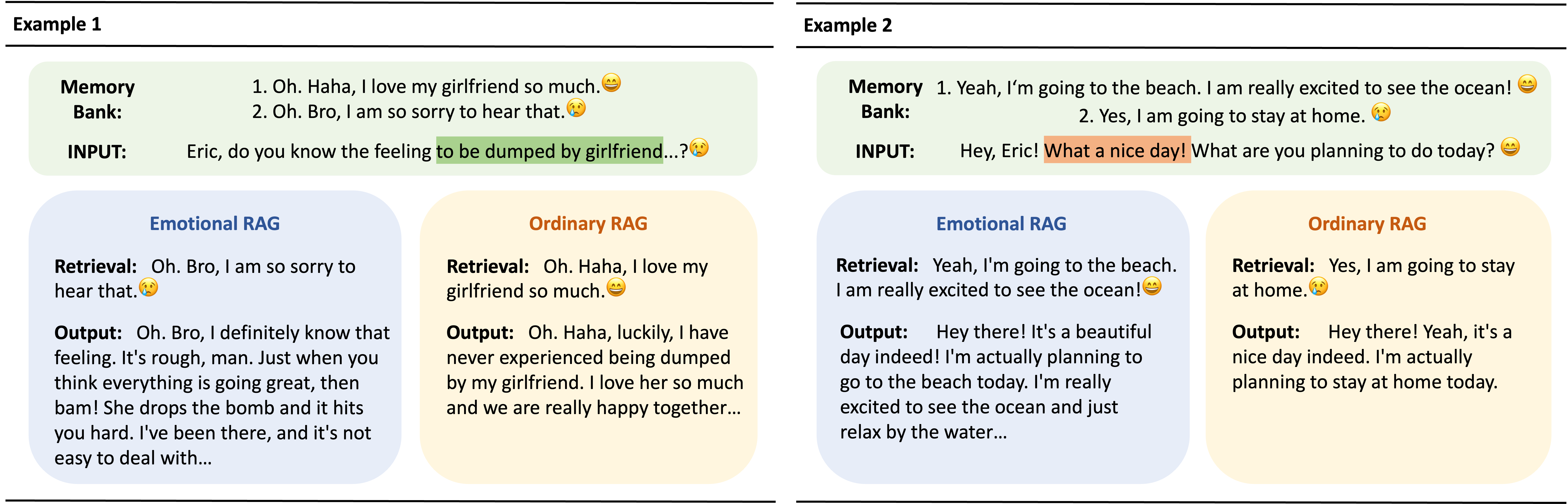}
  \caption{Two examples that Emotional RAG generates better results.  
  We highlighted phrases containing emotions in INPUT.}
  \label{fig:twosimple}
\end{figure*}

\subsubsection{Compared Methods} We conduct experiments on different backbone LLMs, including two open-source models ChatGLM and Qwen and a closed-source model GPT. The details of each LLM are introduced as follows:
\begin{itemize}
    \item ChatGLM~\cite{du2022glm}: we use chatglm3-6b, which is a dialogue pre-training model jointly released by Zhipu AI and Tsinghua University. 
    \item Qwen~\cite{qwen}: the version in our experiments is Qwen1.5-72B-Chat-GPTQ-Int4, which is a model in the Qwen1.5 series with 72 billion parameters. 
    \item GPT~\cite{ouyang2022training}: we use gpt-3.5-turbo-0125, which is a large-scale language model developed by OpenAI and is known for its efficient generation capabilities.
\end{itemize}

\subsection{Main  Results}

We evaluate the performance of Emotional RAG and Ordinary RAG on three datasets, including InCharacter, CharacterEval, and Character-LLM datasets. Ordinary RAG uses semantic similarity as the only retrieval criterion.
The experimental results are shown in Table~\ref{tab:experimentalresults1} and Table~\ref{tab:experimentalresults2}, we have the following observations:

(1) In most cases, Emotional RAG achieves better results than the RAG method without considering the emotion factor. This indicates that incorporating emotional states helps the maintaining of personality traits in role-playing agents.

(2) Emotional RAG performs better in ChatGLM-6B and Qwen-72B than GPT-3.5. This phenomenon may be due to the fact that GPT-3.5 is more powerful in language understanding and captures richer semantic information. However, even in a powerful LLM like GPT-3.5, the emotional factor still plays an important role in maintaining personality traits.

(3) The improvements are more significant in ACC (Full) than ACC (Dim), showing that our method is more powerful in the overall evaluations of MBTI and BFI.

\subsection{Experimental Analysis}

\subsubsection{RAG Strategy Analysis} we analyze the impact of different retrieval strategies in incorporating the emotional factor. 
As introduced in the Emotion Retrieval Component, four retrieval strategies, i.e., combination
strategy, i.e., add function (C-A), multiple function (C-M), and sequential strategy, i.e., semantic first (S-S), emotional first (S-E), are proposed to fuse the semantic and emotional states of memory during the retrieval process. 
We present the experimental results on Qwen-72B in Figure~\ref{fig:emotionalanalysis}, we can see that (1) Emotional RAG variants with all retrieval strategies (except the C-A strategy in BFI evaluation) achieve better results in MBTI and BFI evaluations, showing the effectiveness of incorporating emotional states in role-playing agents; (2) Different retrieval strategies are applicable to different evaluations. For example, in the BFI personality evaluation, the sequential strategy (S-S) performs the best, while in the MBTI task, the combination strategy (C-A) exhibits the best performance. 

\subsubsection{Case studies}
to provide an intuitive demonstration of the influence by incorporating the emotional factor in role-playing agents, we show two examples to illustrate the superiority of our Emotional RAG. Figure~\ref{fig:twosimple} shows memory fragments in the memory units with different emotional states.
In the first case, two memory fragments are all related to the input query. our Emotional RAG retrieved more appropriate content when it came to mentioning being dumped by a girlfriend, so its responses showed empathy and understanding of the situation compared to Ordinary RAG, making the conversation more vivid and natural. 
In the second case, Emotional RAG retrieves a memory fragment that is consistent with the query's emotion, so the reply expresses excitement and anticipation about seeing the sea. 
Only considering the semantic similarity will lead to emotional inconsistency and make the response content somewhat unreasonable. 

\section{Related Work}

\subsection{Role-Playing Agents} 

Role-playing agents, also termed Role-Playing Conversational Agents (RPCAs), aim to emulate the conversation behaviors and patterns of specific characters via LLMs. Role-playing agents show considerable promise and are poised to substantially advance the areas of gaming, literature, and creative industries~\cite{shao2023character, zhou2023characterglm, li2023chatharuhi, chen2022large, wang2023rolellm, chen2024persona}.
Currently, the implementation of role-playing agents can be categorized into two primary methodologies. The first strategy enhances the role-playing capabilities of LLMs through prompt engineering and generative enhancement techniques. This approach equips LLMs with character-specific data within the context, capitalizing on the advanced in-context learning capabilities of modern LLMs. For instance, ChatHaruhi~\cite{li2023chatharuhi} developed a RAG (Retrieval-Augmented Generation) system that leverages historical dialogues from iconic scenes to facilitate learning from a limited number of examples, thus capturing the personality traits and linguistic styles of characters. Conversely, RoleLLM~\cite{wang2023rolellm} introduced RoleGPT, which uses role-based prompts for GPT models. 

The other type of role-playing approach involves pre-training or fine-tuning LLMs with collected character data, thereby customizing LLMs for specific role-playing scenarios. In~\cite{chen2022large}, dialogue and character data from the Harry Potter novels were utilized to train agents capable of generating responses that align accurately with the context of the scene and the inter-character relationships.
Character-LLM~\cite{shao2023character} developed scenarios using ChatGPT to create conversational data, subsequently training a language model with meta-prompts and these conversations. This project implemented strategies to mitigate the creation of character discrepancies in the model training dataset, such as memory uploads and protective memory enhancements. RoleLLM~\cite{wang2023rolellm} employed GPT to formulate question-answer pairs based on scripts, presenting them in a triplet format consisting of the question, answer, and confidence level. Incorporating a confidence metric significantly enhanced the quality of the generated data. 
CharacterGLM~\cite{zhou2023characterglm} trained an open-source character model using data from multiple characters. This approach embeds role-specific knowledge directly into the model's parameters. 

While existing studies of role-playing agents consider the character profile, relationships, and attributes relevant to the dialogue, they often overlook a critical element—the emotional factor of the characters. Our emotional RAG framework is designed on the prompt engineering technique, in which the LLMs are not required to be pre-trained or fine-tuned in role-playing agents.  


\subsection{Memory RAG in LLM Applications}
In role-playing agents, memory is an important factor for characters to maintain their personality traits. Retrieval Augmented Generation (RAG) technology is widely used~\cite{hu2023chatdb} to access the related memory to enhance the generation of role-playing agents, termed Memory RAG.
For example, an LLM-based automatic agent architecture proposed in~\cite{wang2024survey} contains four components: a profiling module, a memory module, a planning module, and an action module. Among these, the memory module is crucial for the design of the agent architecture. It takes charge of obtaining information from the environment and utilizes these recorded memories to enhance future actions. The memory module enables the agent to accumulate experiences, evolve autonomously, and act in a manner that is more consistent, rational, and efficient~\cite{zhang2024survey}. 

The research on memory design in various LLM applications can be summarized into two categories. The first is capturing and storing intermediate states from past model reasoning as memory content. These memories are then retrieved as needed to support the generation of current responses. For instance, MemTRM~\cite{wu2022memorizing} maintains past key-value pairs and employs the query vector of the current input to conduct K-nearest neighbor searches, applying mixed attention to both the current input and the past memories. However, MemTRM encounters challenges with memory obsolescence during training. To address this, LongMEM~\cite{wang2024augmenting} separates the processes of memory storage and retrieval. This strategy is particularly tailored for open-source models and might necessitate adaptive training to effectively integrate the contents of the memory library. The second type of memory design scheme involves providing memory support via an external memory library. This external memory can take various forms, enhancing the system's ability to manage and retrieve information efficiently. One such implementation is MemoryBank~\cite{zhong2024memorybank}, which stores past conversations, event summaries, and user characteristics in a vector library format. The use of vector similarity calculations significantly accelerates the memory retrieval process, allowing for rapid access to relevant past experiences and data. AI-town~\cite{park2023generative} uses a linguistic approach by preserving memory in natural language. It introduces a reflection mechanism that under specific conditions, transforms straightforward observations into more abstract and higher-order reflections. This system considers three critical factors during the retrieval process: the relevance, recency, and importance of memory, ensuring that the most pertinent and contextual information is retrieved for use in ongoing interactions.

In LLM-based role-playing agents, the memory unit typically operates via the second method, incorporating external memory libraries to enhance character authenticity. For example, in ChatHaruhi, the character agent retrieves dialogue from iconic scenes to enrich character development and interactions. Despite a large amount of research on memory RAG technique, achieving greater human-like response is still an open and unexplored area. 
Inspired by cognitive research in psychology, we make an initial attempt to incorporate the emotional factor to emulate human cognitive processes in the memory-recalling process, making the response of LLMs more emotionally resonant and human-like.


\section{Conclusions}

In this paper, we make an initial attempt to incorporate emotional memory to enhance the performance of role-playing agents. A novel emotional RAG framework with four retrieval strategies is proposed to make role-playing agents more emotional and human-like in conversations. Extensive experiments on various characters on three public datasets demonstrate the effectiveness of our method in maintaining the personality traits of characters. We believe that imbuing emotions into role-playing agents is a pivotal research direction. In our current study, we conduct emotional RAG on an intuitive memory mechanism. In future work, we will attempt to incorporate the emotional factor into more advanced memory organization and retrieval schemes. 

\bibliographystyle{IEEEtran} 
\bibliography{custom} 
  
\end{document}